\pdfoutput=1
\documentclass[runningheads]{paper2007-llncs}
\usepackage{times}
\usepackage{helvet}
\usepackage{courier}
\usepackage[misc]{ifsym}
\newcommand\blfootnote[1]{%
  \begingroup
  \renewcommand\thefootnote{}\footnote{#1}%
  \addtocounter{footnote}{-1}%
  \endgroup
}
\usepackage{booktabs}
\usepackage{multirow}
\usepackage{multicol}
\usepackage{amssymb}
\usepackage{subfigure}
\usepackage{hyperref}
\usepackage{array}
\usepackage{comment}
\usepackage{graphicx}
\newcommand{\tabincell}[2]{\begin{tabular}{@{}#1@{}}#2\end{tabular}}  
\begin{document}
\title{Efficient Global-Local Memory for Real-time Instrument Segmentation of Robotic Surgical Video}
\titlerunning{Efficient Global-Local Memory for Instrument Segmentation}

\author{Jiacheng Wang\inst{1},
	Yueming Jin\inst{2},
	Liansheng Wang \inst{1}\textsuperscript{(\Letter)},
	Shuntian Cai\inst{3},\\
	Pheng-Ann Heng\inst{2},
	Jing Qin\inst{4}}

%
\authorrunning{Wang and Jin et al.}
\institute{Department of Computer Science at School of Informatics, Xiamen University \\ \email{jiachengw@stu.xmu.edu.cn,lswang@xmu.edu.cn} \and 
	Department of Computer Science and Engineering,
	The Chinese University of Hong Kong \\
	\email{\{ymjin,pheng\}@cse.cuhk.edu.hk} \and
	Department of Gastroenterology, Zhongshan Hospital affiliated to Xiamen University \\ \email{nktianxingjian@163.com}
	\and Center for Smart Health, School of Nursing, The Hong Kong Polytechnic University \\\email{harry.qin@polyu.edu.hk}}
	
\maketitle              
\begin{abstract}
	Performing a real-time and accurate instrument segmentation from videos is of great significance for improving the performance of robotic-assisted surgery.
	\blfootnote{\textsuperscript{} J. Wang and Y. Jin---Contributed equally.}
	We identify two important clues for surgical instrument perception, including local temporal dependency from adjacent frames and global semantic correlation in long-range duration. However, most existing works perform segmentation purely using visual cues in a single frame. Optical flow is just used to model the motion between only two frames and brings heavy computational cost. We propose a novel dual-memory network (DMNet) to wisely relate both global and local spatio-temporal knowledge to augment the current features, boosting the segmentation performance and retaining the real-time prediction capability. We propose, on the one hand, an efficient local memory by taking the complementary advantages of convolutional LSTM and non-local mechanisms towards the relating reception field. On the other hand, we develop an active global memory to gather the global semantic correlation in long temporal range to current one, in which we gather the most informative frames derived from model uncertainty and frame similarity. We have extensively validated our method on two public benchmark surgical video datasets. Experimental results demonstrate that our method largely outperforms the state-of-the-art works on segmentation accuracy while maintaining a real-time speed\blfootnote{\textsuperscript{} Code is available at \url{https://github.com/jcwang123/DMNet}}.
										
\end{abstract}

\section{Introduction}
Robotic-assisted surgery has greatly improved the surgeon performance and patient safety.
Semantic segmentation of instrument segmentation, aiming to separate instrument and identify its sub-type and parts, serves as an essential prerequisite in various applications in assisted surgery.
Achieving high segmentation accuracy while with low latency for real-time prediction is vital in the real-world deployment.
For example, fast and accurate instrument segmentation can advance the context-awareness of surgeons when performing surgery, providing timely decision-making support, and generating real-time warning of potential deviations and anomalies.
The real-time analysis of robotic surgical videostream can facilitate coordination and communication among the surgical team members.
However, fast and accurate instrument segmentation from surgical video is very challenging, 
due to the complicated surgical scene, various lighting conditions, incomplete and distorted instrument structure caused by small FoV of endoscopic camera and inevitable visual occlusion by blood, smoke or instrument overlap.

Most existing methods~\cite{garcia2017toolnet,winner2017,ni2019attention,ni2020pyramid} on surgical instrument segmentation treat sequential video as static image, and ignore the valuable clues in temporal dimension. For example, the ToolNet~\cite{garcia2017toolnet} uses a holistically-nested fully convolutional network to impose multi-scale constraint of predictions. Shvets \textit{et al.} introduce a skip-connection model trained with transfer
learning, winning the 2017 EndoVis Challenge~\cite{endovis2017}. The LWANet~\cite{ni2019attention} and PAANet~\cite{ni2020pyramid} apply attention-based mechanism into instrument segmentation to encode the semantic dependencies between channels and global space.
Different from these methods that perform segmentation solely relied on visual clues from a single frame, Jin \textit{et al.}~\cite{jin2019incorporating} utilize optical flow to incorporate temporal prior into the network, which is hence able to capture inherent temporal clues from the instrument motion to boost results.
Zhao \textit{et al.}~\cite{zhao2020learning} use motion flow to largely improve the performance of semi-supervised instrument segmentation, with low-frequency annotation available in the surgical video.
However, leveraging optical flow incurs significant computational cost in the calculation process, also can only capture the temporal information in the local range of two frames.

Recently, many works in surgical video analysis have shown successes of using long-range temporal information for improving results, mainly focus on recognition tasks for surgical workflow, gesture and tool presence~\cite{jin2017sv,miccai2019phase,van2020multi,jin2021temporal,czempiel2020tecno,zhang2020symmetric}.
For example, Czempiel et al.~\cite{czempiel2020tecno} proposed to use multi-stage TCN~\cite{farha2019ms} to leverage temporal relations in all previous frame for workflow recognition.
Meanwhile, plenty of studies in natural computer vision domain verify that aggregating the global information that is distant in time to augment the current features can boost the performance.
This series of methods achieves the promising performance on various video analysis tasks, such as action recognition~\cite{wu2019long}, video super resolution~\cite{yi2019progressive}, video object detection~\cite{chen2020memory}, and semi video object segmentation~\cite{voigtlaender2019feelvos,oh2019video}. 
Although these methods are hardly employed in our instrument segmentation task given different problem settings, and generally do not take the low computational latency into consideration which is the core consideration for robotic surgery,
this stream inspires us to raise a question that how to wisely incorporate longer-range cues in temporal dimension for improving both robustness and efficiency to instrument segmentation.

In this paper, we propose a novel Dual-Memory Network (DMNet) for achieving accurate and real-time instrument segmentation from surgical videos, 
by holistically and efficiently aggregating spatio-temporal knowledge. 
The dual-memory framework are based on two important intuitions for humans to perceive instruments in videos, i.e., local temporal dependence and global semantic information, therefore more temporal knowledge can be transferred to current semantic representation.
More importantly, we are the first trials that enable such holistic aggregation in real-time setting by carefully considering the properties of these two-level horizons.
Concretely, we first design a local memory to take the complementary advantages of RNN and self-attention mechanisms.
Considering that the local-range frame segments demonstrate the similar spatial information and highly temporal continuity, we exploit RNN to collect temporal cues with small reception field, following a self-attended non-local block to relate the current frame in larger spatial space with only single collected frame.
We further develop a global memory to enable the current frame to efficiently get access to much longer-span content, by incorporating active learning strategy.
Most informative and representative frames diffusing in global-range segments are selected based on two sample selection criteria.
Relating such frames to current features can achieve sufficient clue enhancement while remaining the fast prediction.
We extensively validate our method on two public benchmark surgical video dataset for instrument segmentation.
Our approach attains a remarkable segmentation improvement over the state-of-the-art methods, with retaining the fast and real-time speed.

\section{Methodology}
\begin{figure*}[t]
	\centering
	\includegraphics[width=1.0\linewidth]{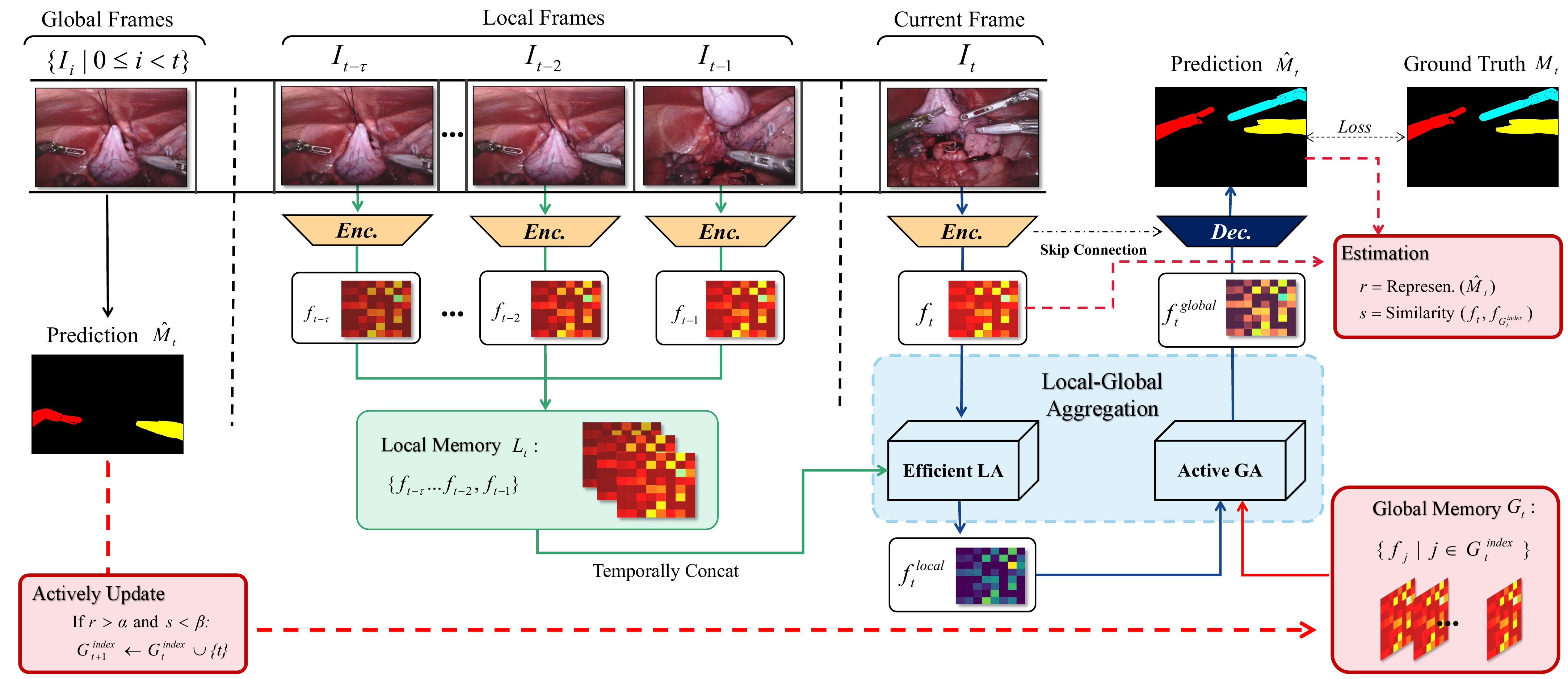}
	\caption{
		Illustration of our proposed Dual-Memory Network. Wisely leveraging both local temporal dependence and global semantic information, our method achieves real-time surgical instrument segmentation with high accuracy.
	}
	\label{fig:frame}
	\vspace{-4mm}
\end{figure*}
Fig.~\ref{fig:frame} illustrates the overall architecture of the proposed dual-memory network (DMNet) for instrument segmentation from surgical videos. Two memories are storing and updating the previous frame knowledge from both local-range and global-range. Efficient local aggregation and active global aggregation modules are designed for relating the two-level memories to enhance the current frame features.

\subsection{Dual-Memory Architecture}
The goal of surgical instrument segmentation is to yield a segmentation map for each frame of the video in real-time to facilitate robot or human manipulation.
Given the current time step as $t$, our dual-memory framework contains a local memory and a global memory for the current frame $I_t$, respectively defined as $\mathbf{L}_t$ and $\mathbf{G}_t$. They are composed of the extracted feature maps $f$ obtained from previous frames, illustrated as the green stream and red stream in Fig.~\ref{fig:frame}, respectively.
The local memory $\mathbf{L}_t$ contains $\tau$ feature maps extracted from previous $\tau$ frames: $\{f_{t-\tau},...,f_{t-2},f_{t-1}\}$,
which is used for local temporal aggregation to avoid redundant computation.
The global memory $\mathbf{G}_t$ is composed of a set of feature maps which are selected from the feature maps of all previous frames.
For global temporal aggregation, we randomly select a certain number of frames from global memory to offer the dependencies, thus will be able to reduce computation cost significantly. 
Leveraging the principle of active learning, 
we further introduce an actively updating strategy to update memory with the most valuable samples to build a more terse global memory.
With local and global aggregation, current frame feature is augmented for accurate instrument segmentation.

\subsection{Efficient Local Temporal Aggregation}
Considering inherent temporal continuity within a video, modeling temporal information, especially from the relevant adjacent frames in a local-range, is crucial to improve segmentation for each frame.
Previous works leverage optical flow to explicitly incorporate the motion information of surgical instrument, yet its calculation is time-consuming.
Convolutional LSTM (\textit{ConvLSTM}) is another powerful approach generally used for temporal modeling.
However, the receptive field captured from each previous frame is limited given the small size of convolutional kernel.
Stacking several layers to increase the receptive field is a relatively common strategy to tackle this problem, yet inevitably brings the large computational cost.
On the other hand, non-local (\textit{NL}) attention mechanism shows its success in image segmentation recently, by enhancing each pixels through weighted consideration on other pixels of all positions.
It is computationally feasible for single image analysis, however, relating several adjacent frames in the video clip shall largely harm the model efficiency.
We propose an efficient local aggregation (ELA) module to take the complementary advantages of ~\textit{ConvLSTM} operation and \textit{NL} mechanism.
Temporal dependency in the local memory can be incorporated using ELA to enhance each frame features with light-weight computation.

Formally, for segmenting the current frame $I_t$, we first form the local feature map clip from the local memory $\{f_{t-\tau},...,f_{t-2},f_{t-1}\}$.
\textit{ConvLSTM} operation is then utilized to aggregate the information along the temporal dimension with small convolution kernel.
Notably, to enable the maximum efficiency, we employ \textit{BottleneckLSTM}~\cite{liu2018mobile} instead of the standard \textit{ConvLSTM}.
It benefits from replacing the standard convolution with depth-wise separable convolution and setting a bottlenecked layer after input to reduce the feature dimension.
We can obtain the temporal enriched feature from the local memory $\tilde{f}_t = \mathcal{F}_{LSTM}(f_{t-\tau},..., f_{t-2}, f_{t-1}, f_{t})$.
However, each pixel in $\tilde{f}_t$ only considers small region information searched within the kernel size.

Instead of stacking the LSTM layers to increase the receptive field, our ELA leverages \textit{NL} mechanism to enlarge the reference region.
Concretely, based on the encoded temporal feature $\tilde{f}_t$ for the current frame, we generate two kinds of feature maps (\textbf{key}, \textbf{value}) of $\tilde{f}_t$, denoted as ($\tilde{\mathbf{k}}_t$, $\tilde{\mathbf{v}}_t$).
Each with its own simple convolution layer that preserves their spatial size, while reducing the dimensionality.
We set the channel number of \textbf{key} feature map as a low value, as we aim to use it to efficiently calculate the similarity matching scores.
While the channel number of \textbf{value} is set as a relatively high value, in order to store more detailed information for producing the mask estimation for surgical instruments.
Next, we compare every spatial location in $\tilde{\mathbf{k}}_t$ with other locations to perform the similarity matching, 
and the similarity function $\mathcal{F}_{sim}$ is defined as:
$\mathcal{F}_{sim}(\mathbf{x},\mathbf{y}) = \exp\left(\mathbf{x} \circ \mathbf{y}\right)$,
where $\circ$ denotes the dot product. 
The value in $\tilde{\mathbf{v}}_t$ is then retrieved by a weighted summation with
the soft weights and concatenated with the original $\tilde{\mathbf{v}}_t$.
To this end, output feature calculated by our ELA module is as follow:
\begin{eqnarray}\label{eq:relation}
	\vspace{-2mm}
	{f}_t^{local_i} = \mathcal{F}_{NL}(\tilde{f}_t^i) = [ \tilde{\mathbf{v}}_t^i, 
	\frac{1}{\mathcal{Z}} \sum_{\forall j} \mathcal{F}_{sim}(\tilde{\mathbf{k}}_t^i,\tilde{\mathbf{k}}_t^j)\tilde{\mathbf{v}}_t^j],
	\vspace{-2mm}
\end{eqnarray}
where $\mathcal{Z} = \sum_{\forall j} \mathcal{F}_{sim}(\tilde{\mathbf{k}}_t^i,\tilde{\mathbf{k}}_t^j) $ is the normalizing factor; $[\cdot,\cdot]$ denotes the concatenation;
$i$ is the index of an output position whose response is to be computed and $j$ is the index that enumerates all possible positions in the temporal enriched feature $\tilde{f}_t$. 
In this regard, ${f}_t$ is enhanced to ${f}_t^{local}$ by the efficient feature aggregation along both temporal and visual space from our built local memory.

\subsection{Active Global Temporal Aggregation}
Apart from augmenting the current frame feature using the local temporal dependency, we propose to introduce the cues that is more distant in time from the global memory.
The standard approach for global temporal aggregation is to put feature maps of all the previous frames into the global memory and randomly select some samples when activating it.
However, this approach has significant memory-cost, particularly when the video is long and contains plenty of frames. 
More importantly, different frames indeed count for different values for global temporal aggregation.
For example, frames with the motion blur and lighting reflection contain little beneficial information.
Including them into the global memory shall contribute less to augment the current frame features.
Additionally, adjacent frames commonly have similar visual appearance, leading to redundant features which help nothing but cost more.
We propose to selectively recommend frames to form a more supportive global memory $\mathbf{G}_t$ for aggregation, named as active global aggregation (AGA).
We introduce two selection criteria complementarily considering both representativeness and similarity when forming the global memory, where the representativeness denotes the general semantic context and the similarity is used to evaluate whether the coming frame is supplementary or redundant.

Specifically, given the segmentation map $\hat{M}_t$ of the $t$-th frame,
we first compute the prediction confidences $\mathbf{r}$ through the entropy on behalf of the representativeness.
As the samples with low confidence are more likely to be abnormal, causing the prediction at decision boundary, we only queue the frame into $\mathbf{G}_t$ if $\mathbf{r}>\alpha$.
The predicted entropy is defined as:
$\mathbf{r} = \frac{1}{N} \sum_{i} \sum_c p^c_i \log p^c_i$,
where $p^c_i$ denotes probability of the $c$-th class at location $i$ and $N$ is the number of pixels in the map.

We also aim at the sample diversity in the global memory.
The straightforward way is to make a comparison with all feature maps when adding a new one.
However, the full comparison is computation-intensive.
Instead, we only compare with the latest queued feature map $f_{\mathbf{G}_{t}^{latest}}$ in the global memory $\mathbf{G}_{t}$.
If dissimilar, the newly coming one is less likely to resemble further features.
We employ a negative euclidean distance to measure the multivariate similarity of a pair of feature maps as $\mathbf{s} = -\sqrt{\sum(f_t-f_{\mathbf{G}_{t}^{latest}})^2}$.
We include the feature map into the global memory if $\mathbf{s}<\beta$.

After we build the global memory $\mathbf{G}_t$ in which each element is generally informative, we randomly select $n$ frame features $\{f_{g}\}_{g=1}^n$, from $\mathbf{G}_t$ for aggregation.
The AGA module is also developed based on the \emph{NL} mechanism.
However, considering that the $n$ frame features are independent without holding temporal continuity, AGA module separately relates each of them to enhance current frame feature.
Therefore, different from ELA module to conduct self-attention of feature itself, AGA module input two kinds of features including $\{f_{g}\}$ and the current frame features augmented by ELA $f_t^{local}$.
Two pairs of (\textbf{key}, \textbf{value}) are calculated for different features, while other operations such as similarity matching remain the same as ELA module.
We denote the globally aggregated feature as $f^{global} = \mathcal{F}_{NL}(f_t^{local}, \{f_{g}\}_{g=1}^n)$.
To this end, the $f^{global}$ can capture the semantic content of the frames that is distant in time.
Note that, to ensure the online prediction, only previous frame features are accessible in global memory for augmenting each frame feature.
The $f^{global}$ are then decoded to produce the final mask of surgical instrument.
We employ \textbf{Dice} loss between ground-truth and predicted mask for model optimization.

\section{Experiments}
\textbf{Datasets.}
We extensively evaluate the proposed model on two popular public datasets: 2017 MICCAI EndoVis Instrument Challenge (EndoVis17) and 2018 MICCAI EndoVis Scene Segmentation Challenge (EndoVis18). 
EndoVis17~\cite{endovis2017} records different porcine procedures using da Vinci Xi surgical system.
For fair comparison, we follow the same evaluation manner in ~\cite{winner2017,jin2019incorporating}, by using the released 8$\times$ 225-frame videos for 4-fold cross-validation, also with the same fold division.
We perform the most challenging problem, type segmentation, on this dataset to validate our method.
EndoVis18~\cite{allan20202018} made up of 19 more complex surgery sequences with 15 training videos and 4 test videos. We randomly split three videos out of training sets for model validation. 
Complete annotation in this challenge contains 12 classes which also includes anatomical objects.
Among them, surgical instruments are labeled by different parts (shaft, wrist, and jaws). In this work we focus on these three classes regarding instruments and ignore other segmentation classes.
Two commonly used evaluation metrics Mean intersection-over-union (mIoU) and mean Dice coefficient (mDice) are adopted for validation.
To evaluate model complexity and time performance, we calculate the number of parameters (Param.), FLOPS, inference time and FPS.
We average the inference time on a single frame by 100 times for fair comparison.
\\
\textbf{Implementation Details.} 
We adopt the lightweight \emph{RefineNet}~\cite{nekrasov2018light} as the backbone and replace its encoder with \emph{MobileNetv2}~\cite{Sandler_2018_CVPR}. 
All the training images are resized to $512$ $\times$ $640$ and augmented by vertical flip, horizontal flip and random scale change (limited 0.9-1.1). 
The network is pre-trained on ImageNet~\cite{imagenet_cvpr09}, and optimized by Adam~\cite{kingma2017adam}, with weight decay as 0.0001.
We take a mini-batch size of $16$ on $4$ TITAN RTX GPUs and apply synchronous Batch Normalization to adapt multi-GPU environment~\cite{Peng_2018_CVPR}.
Learning rate is initialized as 0.0001 and reduced by one-tenth in the 70-th epoch with training 100 epochs in total.
We empirically set $\tau=4$ and $n=4$ considering the trade-off between accuracy and efficiency. 
We set $\alpha = -0.08$ and $\beta=-4.65$, by averaging $\mathbf{r},\mathbf{s}$ of all frames, to represent the general distribution.
All experiments are repeated 5 times and we report average values to account for the stochastic nature of training.
\\
\textbf{Comparison with State-of-the-arts.}
We compare the proposed DMNet with state-of-the-art approaches, including: 
(1) Ternausnet~\cite{ternausnet}: the winner method of EndoVis17,
(2) MF-TAPNet~\cite{jin2019incorporating}: the latest instrument segmentation model that applies temporal information by optical flow,
(3) PAANet~\cite{ni2020pyramid}: a 2D segmentation model achieving best performance in instrument segmentation,
(4) LWANet~\cite{ni2019attention}: the latest instrument segmentation model with real-time performance.
(5) TDNet~\cite{hu2020temporally}: a NL-based model processing local-range temporal context in real-time for natural video analysis.
For the first two methods, we refer the results from their papers.
We re-implement PAANet, LWANet, TDNet and perform 4-fold cross-validation for fair comparison. 
\begin{table*}[t]
	\caption{Type and part segmentation results of instrument on the EndoVis17 and EndoVis18 datasets, respectively. 
		Note that underline denotes the methods in real-time.
	}
	\vspace{-5mm}
	\label{table:results2017}
	\begin{center}
		\renewcommand\arraystretch{1.2}
		\begin{tabular}{c||cc|cc|c|c|c|c}
			\hline
			\hline
			\multirow{2}{*}{\bf Methods} & \multicolumn{2}{c|}{\textbf{EndoVis17}} &
			\multicolumn{2}{c|}{\textbf{EndoVis18}} &
			\multirow{2}{*}{\small \bf \tabincell{c}{Param.\\(M)}} &
			\multirow{2}{*}{\small \bf \tabincell{c}{FLOPS\\(G)}} &
			\multirow{2}{*}{\small \bf \tabincell{c}{Time\\(ms)}} &
			\multirow{2}{*}{\small \bf FPS} \\
			&\scriptsize \bf mDice(\%) &\scriptsize \bf mIOU(\%) &
			\scriptsize \bf mDice(\%) &\scriptsize \bf mIOU(\%) &&&&\\
			\hline
			TernausNet~\cite{ternausnet}          & 44.95          & 33.78          & 61.78          & 50.25          & 36.92 & 275.45 & 58.32 & 17             \\
			MF-TAPNet~\cite{jin2019incorporating} & 48.01          & 36.62          & -              & -              & -     & -      & -     & -              \\
			PAANet~\cite{ni2020pyramid}           & 56.43          & 49.64          & 75.01          & 64.88          & 21.86 & 60.34  & 38.20 & 26             \\
			LWANet~\cite{ni2019attention}     & 49.79 & 43.23 & 71.73 & 61.06 &
			\textbf{2.25} & \textbf{2.77} & 13.21 & \underline{76} \\
																														        
			TDNet~\cite{hu2020temporally}         & 54.64          & 49.24          & 76.22          & 66.30          & 21.23 & 47.60  & 22.23 & \underline{45} \\
			\hline
			DMNet                                 & \textbf{61.03} & \textbf{53.89} & \textbf{77.53} & \textbf{67.50} & 4.38  & 11.53  & 26.37 & \underline{38} \\
			\hline
		\end{tabular}
	\end{center}
	\vspace{-5mm}
\end{table*}

\begin{figure}[t]
	\centering
	\includegraphics[width=0.85\linewidth]{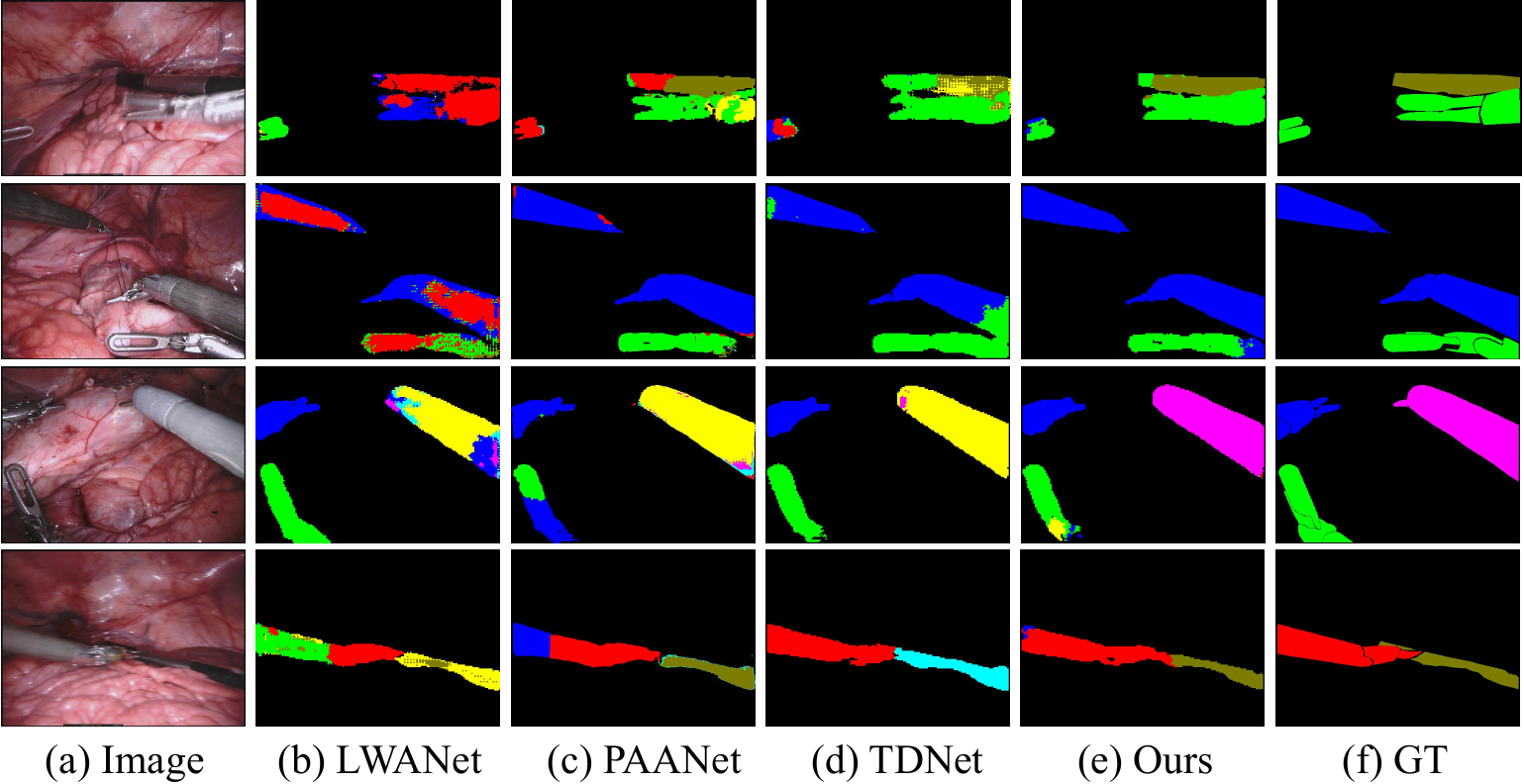}
	\vspace{-2mm}
	\caption{Visual comparison of type segmentation on EndoVis17 produced by different methods.}
	\label{fig:comparison_result}
	\vspace{-6mm}
\end{figure}
As shown in Table~\ref{table:results2017}, DMNet achieves 61.03\% mDice and 53.89\% mIoU on instrument type segmentation, outperforming the other methods in terms of both metrics by a large margin.
Without using temporal aggregation, the real-time approach LWANet only achieves the 49.79\% on mDice and 43.23\% on mIoU, even if it's the fastest model, demonstrating the importance of temporal information in boosting the segmentation performance.
In particular, compared with TDNet, which also harnessed local temporal context to improve performance, the proposed DMNet yields better segmentation accuracy with a nearly closed speed, demonstrating the effectiveness of the proposed active global aggregation scheme.
What's more, our results achieve the consistent improvement in part segmentation on EndoVis18 dataset, outperforming our rivals on the test set.
It proves that our local-global temporal context is not limited to type segmentation, but can provide general feature enhancement.
We further present visual comparison results of some challenging cases in Fig.~\ref{fig:comparison_result}.
It is observed that without temporal information, models can hardly identify different types (Fig.~\ref{fig:comparison_result} (b,c)). 
Solely using local (Fig.~\ref{fig:comparison_result} (d)) temporal information, TDNet still performs unsatisfactory segmentations with structural incompleteness caused by pose variation (3rd row) and large occlusion caused by tool overlapping (4th row).
Our DMNet achieves segmentation results closest to the ground truth, demonstrating the effectiveness of the proposed temporal information aggregation schemes in tackling challenging cases.
\\
\textbf{In-depth Analysis.}
\begin{table}[!t]
	\caption{In-depth analysis of our method.}
	\centering
	\vspace{-3mm}
	\subtable[Ablation for key components.]{
		\begin{tabular}{c|c||c|c}
			\hline
			\hline
			\bf ELA    & \bf AGA    & \bf mIoU(\%) & \bf FLOPS(G) \\
			\hline
			           &            & 46.51        & 10.21        \\
			\checkmark &            & 52.25        & 10.93        \\
			           & \checkmark & 51.64        & 10.77        \\
			\hline
			\checkmark & \checkmark & 53.89        & 11.53        \\
			\hline
		\end{tabular}
		\label{table:ablation-study}
	}
	\qquad
	\subtable[Different fashions for local aggregation.]{
		\begin{tabular}{l||c|c|c}
			\hline
			\hline
			\bf Type  & \bf mIoU(\%) & \bf Param.(M) & \bf FLOPS(G) \\
			\hline
			NL        & 48.03        & 6.40          & 11.82        \\
			CLSTM     & 53.07        & 7.93          & 17.68        \\
			BLSTM     & 50.14        & 3.65          & 10.84        \\
			\hline
			ELA(Ours) & 52.25        & 4.05          & 10.93        \\
			\hline
		\end{tabular}
		\label{table:type}
	}
	\vspace{-2mm}
\end{table}
\begin{figure}[!t]
	\centering
	\includegraphics[width=1.0\linewidth]{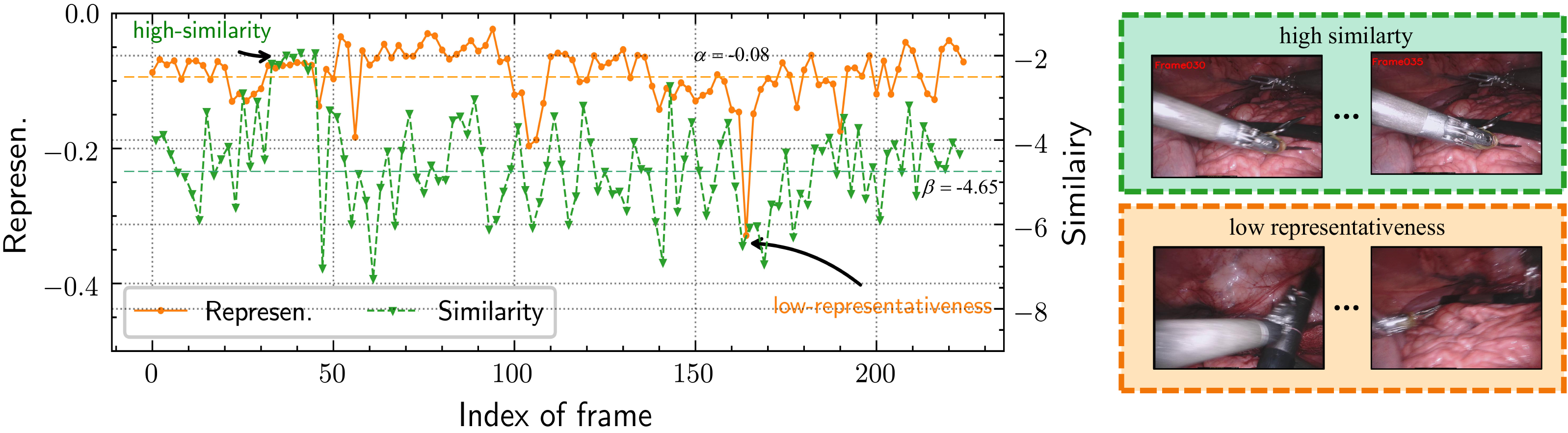}
	\vspace{-4mm}
	\caption{Similarity and representativeness of a video sequence in global aggregation.}
	\label{fig:sim_rep}
	\vspace{-5mm}
\end{figure}
We analyze effectiveness of each key component proposed in our method on EndoVis17. 
As shown in Table~\ref{table:ablation-study}, applying ELA individually increases $5.74\%$ of mIoU, 
while adopting AGA individually improves $5.13\%$ of mIoU. 
With both ELA and AGA, mIoU further boosts by $7.38\%$,
indicating the complementary advantages of local and global information aggregation.
More importantly, we achieve these improvements with comparable FLOPS.
Without introducing much computational cost, our method maintain real-time inference capability.

We conduct an analysis on our ELA module design, by comparing it with other aggregation manners, including 
(1) a \textit{NL} operator, which is a basic aggregation module attempting to leveraging more spatial information and  
(2) two \textit{ConvLSTM} operators, a standard convolutional LSTM (CLSTM) and a Bottleneck LSTM (BLSTM).
As shown in Table~\ref{table:type}, our ELA module outperforms NL and BLSTM in mIoU by $4.18\%$ and $2.11\%$, respectively, with little extra computational cost.
While CLSTM achieves slightly better performance than ELA, its computational costs are almost twice more than ELA.
The experimental results demonstrate the ELA achieves a good balance between accuracy and efficiency.  

We present the calculated $\mathbf{r}$ and $\mathbf{s}$ of a video sequence and visualize some typical frame samples in Fig.~\ref{fig:sim_rep}, to verify the motivation of active selection in proposed AGA module.
We see from the green box that the video sequence with close similarity scores shows high-similar appearance. 
In the orange box, we see that the sample with low confidence presents the abnormal appearance due to motion blur.
Using the global information of data, i.e., average values of $\alpha$ and $\beta$ as thresholds, our AGA module bypasses to utilize such information.

\section{Conclusion}
This paper presents a novel real-time surgical instruments segmentation model by efficiently and holistically considering the spatio-temporal knowledge in videos.
We develop an efficient local cache for leveraging the most favorable region per frame for local-range aggregation, and an active global cache to select the most informative frames to cover the global cues using only few frames.
Experimental results on two public datasets shows that our method outperforms state-of-the-arts by a large margin in accuracy while maintaining the fast prediction speed.

\bibliographystyle{paper2007}
\bibliography{paper2007}

\end{document}